\documentclass{article}

\usepackage{nac_preprint}

\usepackage{times}
\usepackage{soul}
\usepackage{url}
\usepackage[hidelinks]{hyperref}
\usepackage[utf8]{inputenc}
\usepackage[small]{caption}
\usepackage{graphicx}
\usepackage{amsmath}
\usepackage{amssymb}
\usepackage{amsthm}
\usepackage{booktabs}
\usepackage{algorithm}
\usepackage{algorithmic}
\usepackage[switch]{lineno}
\usepackage{comment}

\usepackage{xcolor}

\newcommand{\specialcell}[3][c]{%
  \begin{tabular}[#1]{@{}#2@{}}#3\end{tabular}
  }

\makeatletter
\renewcommand{\@makefnmark}{\makebox{\normalfont[\@thefnmark]}}
\makeatother

\title{A Review of Neuroscience-Inspired \\Machine Learning}

\author{%
  Alexander Ororbia \\
  Rochester Institute of Technology \\
  Rochester, NY 14623, USA \\
  \texttt{ago@cs.rit.edu}
  \And 
  Ankur Mali \\
  University of South Florida \\
  Tampa, FL 33620, USA \\
  \texttt{ankurarjunmali@usf.edu}
  \And 
  Adam Kohan \\
  University of Massachusetts Amherst \\
  Amherst, MA, USA \\
  \texttt{akohan@umass.edu}
  \And 
  Beren Millidge \\
  Zyphra, Palo Alto, CA \\
  University of Oxford, Oxford, UK\\
  \texttt{beren@millidge.name}
  \And 
  Tommaso Salvatori \\
  VERSES AI Research Lab, Los Angeles, USA \\
  TU Wien, Vienna, Austria \\
  \texttt{tommaso.salvatori@verses.ai}
}

\begin{document}

\maketitle
\begin{abstract}
    One major criticism of deep learning centers around the biological implausibility of the \emph{credit assignment} schema used for learning -- backpropagation of errors. This implausibility translates into practical limitations, spanning scientific fields, including incompatibility with hardware and non-differentiable implementations, thus leading to expensive energy requirements. In contrast, biologically plausible credit assignment is compatible with practically any learning condition and is energy-efficient. As a result, it accommodates hardware and scientific modeling, e.g. learning with physical systems and non-differentiable behavior. Furthermore, it can lead to the development of real-time, adaptive neuromorphic processing systems. In addressing this problem, an interdisciplinary branch of artificial intelligence research that lies at the intersection of neuroscience, cognitive science, and machine learning has emerged. In this paper, we survey several vital algorithms that model bio-plausible rules of credit assignment in artificial neural networks, discussing the solutions they provide for different scientific fields as well as their advantages on CPUs, GPUs, and novel implementations of neuromorphic hardware. We conclude by discussing the future challenges that will need to be addressed in order to make such algorithms more useful in practical applications.

\end{abstract}

\section{The Problem of Credit Assignment}
\label{sec:intro}


One of the key tasks in artificial intelligence is to construct mathematical and algorithmic solutions to what is known as the grand problem of \emph{credit assignment}. Effective credit assignment reduces to: (i) the identification of which neural processing elements (NPEs), e.g., individual computational units in a computation graph, have an influence on a particular (task-specific) objective functional $\mathcal{L}(\Theta)$; and (ii) modifying the synapses that connect all of the NPEs based on their degree of influence so as to optimize this objective. The synaptic adjustments that characterize the second step are made to improve the overall performance of the network that the set of NPEs constitutes. From the perspective of error-driven learning and adaptation, credit assignment is typically carried out by computing and assigning error values to each NPE based on the cost $\mathcal{L}(\Theta)$ and, once these values have been obtained, yielding $\Delta$ (the set of all adjustments to be made to the synapses within $\Theta$), the current values of the ANN's parameters are consequently updated. 

Note that, historically, error-driven adjustment of the kind described above has been theorized \cite{rao1999predictive,friston2005theory} and experimentally observed in biological neuronal networks \cite{summerfield2006predictive}. The problem with how this error is computed and allocated in modern-day ANNs -- via backpropagation of errors \cite{linnainmaa1970representation} (backprop) --  is essentially what is considered to be neurobiologically implausible. 
%
Given the significant successes of modern deep learning, addressing this implausibility may seem a niche task of interest to neuroscientists alone. However, this is far from true; despite recent breakthroughs, important advancements are still required. Two of these include the need to: 
\textbf{1)} develop more robust and more general human-like capabilities, in service of the grander goal of constructing intelligent machines, and 
\textbf{2)} to construct faster, more energy efficient procedures for training and conducting inference in such models. Biologically-plausible credit assignment approaches offer a potential solution to both of these problems. 

Biologically-plausible (bio-plausible) credit assignment is suitable for neuromorphic hardware implementations due to the \emph{locality} of their operations and synaptic updates. When translated to hardware, locality enables the full parallelization of operations, with low latency, low power consumption, and often without supervision. This is different from processing in existing Von-Neumann architectures, where the division between memory and computing units makes such operations slower and computationally expensive. Locality of operations is also a key property that enables the training of networks with cyclic and entangled topologies, like those of neural circuits, without the need to save gradients in memory, as is required by backprop through time \cite{salvatori2022learning,kohan2023signal,salvatori2023causal}. 


To develop such neuroscience-inspired algorithms, researchers have looked to the neuronal cells that make up animal and human brains, crafting approaches that conduct `credit assignment' \cite{bengio1993credit} in a fashion more analogous to neurobiological dynamics and information processing. Historically, these efforts have drawn from insights and findings in neuroscience, cognitive science, and biophysics. Today, a large amount of this research focuses on developing energy-based, forward only, or spiking algorithms that are stable and perform reasonably well on standard deep learning tasks \cite{scellier2023energy,salvatori2023brain,ernoult2022,kohan2023signal}. The aim of this work is to survey these algorithms and procedures.


\paragraph{Organization of the Review.} This survey is organized in the following manner: In Section 1, we briefly describe several of the key criticisms of credit assignment conducted via backpropagation of errors (BP); In Section 2, we turn to the examination of several emergent paradigms in the realm of neuroscience-oriented algorithms for credit assignment; In Section 3, we discuss potential of neuromprphic systems. In light of the learning and credit assignment schemes studied, in Section 4, we consider important open questions and challenges facing research in neuroscience-inspired machine learning as well as promising problem domains in which advances might be made. Finally, we conclude this targeted survey with final remarks.

\subsection{What is Wrong with Backprop?} 

Despite its impressive empirical successes, BP-based credit assignment is still considered biologically implausible in the brain. We next discuss some of the major criticisms, while the rest of the survey will focus on neuroscience-inspired algorithms that ameliorate some of these issues.

\noindent
\textbf{Weight Transport (WT).}  This term refers to models that use the same set of weights to perform both the forward and backward passes. In neural networks trained with backprop, presynaptic NPEs receive error gradient information from postsynaptic ones via the same synaptic connections that were used to originally forward propagate information. This operation is implausible in the chemical synapses of the brain, where unidirectional flow of information is enforced by neurotransmitters and receptors. In neuromorphic chips, where physical components are used to emulate biological synapses, implementing bi-directional connections may be problematic, depending on the considered hardware \cite{lillicrap2016random,whittington2019theories}.

\noindent
\textbf{Forward Locking (FL).} This term refers to the requirement that the neural activities in one layer of a network cannot be computed until the activities of all preceding layers have been computed. This sequential dependency creates a bottleneck in information processing and is a departure from the parallel, distributed nature of computation in biological networks \cite{jaderberg2016decoupled}. Furthermore, this requires storage of the neurons' values in memory, which is challenging to implement on local and parallel neuromorphic hardware.

\noindent
\textbf{Backward Locking (BL).} Similarly, update/backward locking refers to the delay in computing teaching signals and synaptic updates for a  layer until the teaching signals in the subsequent layers have been computed. This dependency is also at odds with the local, parallel processing observed in biological neural systems \cite{jaderberg2016decoupled}.

\noindent
\textbf{Forward-Backward Differentiation (FBD).} In backprop, the forward and backward passes utilize different computations. The forward pass transmits information across the network, while the backward pass produces gradient information. This divergence in computation between the two passes is seen as implausible and contrasts sharply with the localized and time-constrained plasticity of real synaptic connections \cite{whittington2019theories,lillicrap2020backprop}.

\begin{table}[!t]
\small{
\begin{center}
\begin{tabular}{|r|c|} 
\hline
\textbf{Term} & \textbf{Definition}\\ [0.5ex] 
\hline\hline
PC & Predictive Coding\\
FL & Feedback Alignment\\
BL & Backpropagation Learning\\
FO & Forward Only Learning\\
LRA & local representation learning\\
\hline
FL & Forward Locking\\
BL & Backward Locking\\
WT & Weight Transport\\
FBD & Forward Backward Computation\\
\hline
Parallel & Parallel learning\\
Async. & Asynchronous inference and Learning \\
1-compute & No divergence in computation \\
No-diff & Differentiation not required\\
Sparsity & Sparse learning signals and architecture\\

\hline
\end{tabular}
\end{center}
\caption{Terms/acronyms reference table.}
\label{table:terms}
}
\vspace{-0.3cm}
\end{table}

\section{Neuroscience-Inspired Credit Assignment}
\label{sec:bio_ca}

In this section, we examine several prominent and promising credit assignment paradigms, that have recently garnered a growing body of theoretical and empirical support. To study these credit assignment processes, we draw inspiration from \cite{millidge2022backpropagation}, and reformulate the most general form of each scheme in light of what compound global energy functional that it can be stated to be optimizing. In most of the algorithms considered here, the energy functional will be divided in two terms, $\mathcal{L}(\Theta_L)$, defined on the output layer and related to the objective of the specific task, and $\mathcal{E}(\Theta)$ (note that $\Theta = \Theta_L \cup \{\Theta_\ell\}^{L-1}_{\ell=1}$ where $\Theta_\ell$ contains parameters for layer $\ell$), related to the internal energy of the model, that allows learning via local messages: 
\begin{align}
    \mathcal{F}(\Theta) &=  \beta \mathcal{L}(\Theta_L) + \alpha \mathcal{E}(\Theta).
\end{align}
\paragraph{Notation.} The key notation and symbols that will be commonly used throughout this article is presented here. A bold-font capital character, e.g., $\mathbf{M}$, is used to represent a matrix while a lowercase bold one, e.g., $\mathbf{v}$, represents a vector. $M_{ij}$ effectively retrieves a scalar at position $(i,j)$. In terms of operations, matrix-matrix/vector multiplication is shown as $\cdot$, a Hadamard product is denoted by $\odot$, and $(\mathbf{v})^{\mathsf{T}}$ indicates the transpose of $\mathbf{v}$. An elementwise function, e.g., an activation function, will typically be represented by $\phi(\mathbf{v})$ and $\partial \phi(\mathbf{v})$ is its first derivative with respect to its input $\mathbf{v}$. Generic functions are represented with an italicized letter with a subscript denoting what parameters or subset of parameters it depends on, e.g., ${f}_\Theta()$ which depends on parameter values inside the model parameter construct $\Theta$.

\subsection{Backpropagation of Errors}
\label{sec:backprop}

The general form of the objective that backpropagation (backprop) is used to optimize can be extracted from an energy functional with the form depicted as follows:
\begin{align}
    \mathcal{F}(\Theta) = \beta \mathcal{L}(\Theta)\Big|_{\beta=1} + \alpha \mathcal{E}(\Theta)\Big|_{\alpha=0} = \mathcal{L}(\Theta), \label{eqn:backprop_energy}
\end{align}
where $\alpha$ and $\beta$ are sensitivity hyperparameters/coefficient. Specifically, for backprop, we see that the coefficient which weights the internal local objectives -- $\mathcal{E}(\Theta)$ -- is set to zero ($\alpha = 0$), reducing optimization to exclusively using the task-centric cost.

\paragraph{Learning Dynamics.} To optimize Equation \ref{eqn:backprop_energy}, reverse-mode differentiation is used to calculate the partial derivatives of $\mathcal{F}(\Theta)$ with respect to $\Theta$, which are then subsequently used to carry out a step of gradient descent. This reverse-mode differentiation consists of two operations or phases: a forward pass, that assigns values to every neuron of the network, and a backward pass, that performs credit assignment by backpropagating information about the gradients of $\mathcal{L}(\Theta)$, one layer at the time. The backward pass is performed using the same weight matrices of the forward pass, causing the WT problem described earlier. Furthermore, the sequentiality of the three phases -- forward pass, backward pass, and weight update -- forces the saving of the values computed during each phase before starting the following one, causing the problems of FL and BL.



\subsection{Predictive Coding} 
\label{sec:predictive_coding}

\paragraph{Background.} Predictive coding is a general theory of cortical function which draws inspiration from computational neuroscience, signal processing, and Bayesian inference \cite{friston2003learning,rao1999predictive,salvatori2023brain,song2024inferring}. Predictive coding (or PC) argues that the fundamental principle underlying cortical computation is that of minimizing prediction error: neurons predict their input and output neurons -- typically in a layerwise hierarchy which means that neurons predict the activities of the layer below them. Information about the discrepancies in these predictions are relayed upwards through the hierarchy, serving as the basis for parameter adjustment. In machine learning, recent studies have explored the efficacy of this algorithm across a diverse array of tasks and problem domains, encompassing areas like computer vision, natural language processing, graph learning, and associative memories \cite{millidge2022predictive,pinchetti2022predictive,tang2023recurrent,ororbia2022ngc,salvatori2021associative} where predictive coding networks have been shown to perform on par with backpropagation at training complex machine learning architectures. PC has also been shown, in certain limits, to closely approximate backpropagation of errors \cite{whittington2019theories,millidge2022backpropagation,song2020can,salvatori2022reverse}.

\paragraph{Learning Dynamics.} 
The energy function underlying predictive coding circuitry is as follows:
\begin{align}
    \mathcal{E}(\Theta) &= \sum^{L-1}_{\ell=1} \mathcal{C}(\mathbf{z}^\ell, f_{\Theta_\ell}(\mathbf{z}^{\ell-1}), \mathbf{\Sigma}^\ell) \label{eqn:pc_energy}
\end{align}
where the local measurement takes the following form:
\begin{align}
    \mathcal{C}(\mathbf{z}^\ell, f_{\Theta_\ell}(\mathbf{z}^{\ell-1}), \mathbf{\Sigma}^\ell) = - \frac{1}{2 \Sigma^\ell}\Big|\Big| \mathbf{z}^\ell - f_{\Theta_\ell}(\mathbf{z}^{\ell-1})\Big|\Big|^2_2
\end{align}
noting that $\Sigma^\ell$ is a scalar covariance weighting associated with layer $\ell$ and is typically treated as part of $\Theta_\ell$. Note that, in some variations of PC, $\Sigma^\ell$ is a learned covariance matrix (and its inverse is referred to as the precision); we do not treat this form here and refer the reader to \cite{salvatori2023brain} for details.
 
To optimize the functional in Equation \ref{eqn:pc_energy}, gradients are taken, for each layer, with respect to both the neural activity -- $\frac{\partial \mathcal{E}(\Theta)}{\partial \mathbf{z}^\ell}$ -- and the parameters --  $\frac{\partial \mathcal{E}(\Theta)}{\partial \Theta_\ell}$ -- to conduct a form of expectation maximization (EM) \cite{dempster1977maximum}. For all layers $\ell = 1,...,L$, neural activity values, initialized to base condition values, are iteratively adjusted via a form of $\mathbf{z}^\ell \leftarrow \mathbf{z}^\ell - \gamma \frac{\partial \mathcal{E}(\Theta)}{\partial \mathbf{z}^\ell}$ for $T$ steps (the E-step). Once activities have been updated, each layer's parameters are updated with one application of $\Theta_\ell \leftarrow \Theta_\ell - \gamma \frac{\partial \mathcal{E}(\Theta)}{\partial \Theta_\ell}$ (M-step). Although the optimization of a PC circuit generally follows the gradient flow induced by the free energy functional of Equation \ref{eqn:pc_energy}, the updates to local parameters $\Theta_\ell$ can be written down as multi-factor Hebbian rules. Furthermore, various research efforts have demonstrated that the message passing of error information between layers can be done with decoupled feedforward and feedback synaptic pathways \cite{ororbia2022ngc,salvatori2023brain}.

\begin{figure*}[!t] 
\centering  
\includegraphics[width=1.0\linewidth]{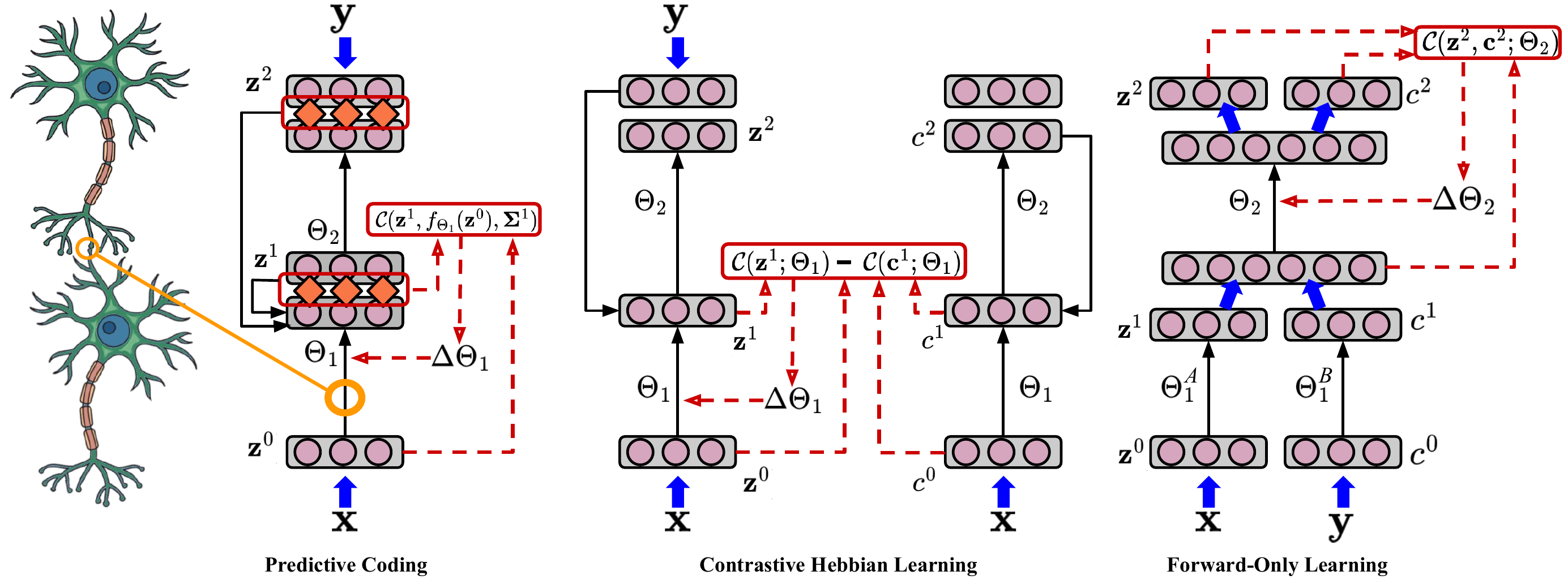}
\caption{
Visualization of the mechanics inherent to predictive coding (PC), contrastive Hebbian learning (CHL), and forward-only learning (specifically signal propagation) when conducting supervised learning. A reddish-orange diamond represents an error neuron (a mismatch signal) while a pinkish circle represents an NPE. Black solid arrows represent synaptic pathways, thicker blue arrows represent non-synaptic transmission pathways (e.g., copying a vector, concatenation, vector-splitting, etc.), and red-dashed arrows just represent the inputs that enter a local energy function. Note that some visualized algorithms (PC, CHL) entail an iterative settling process that is not depicted.
}
\label{fig:bio_algos}
\end{figure*}

\subsection{Contrastive Hebbian Learning}
\label{sec:contrastive_hebb}

\paragraph{Background.}  Schemes that fall under the framework of contrastive Hebbian learning (CHL) \cite{xie2003equivalence}, directly utilize an energy based model that relaxes to solutions. In CHL algorithms such as equilibrium propagation (EP) \cite{scellier2018generalization}, two different phases are required to perform credit assignment: a prediction/unclamped phase, where energy is particularly minimized in order to make a prediction based only on the input data, and a learning/clamped phase, where the output neurons are nudged towards a supervised signal. Once both phases are complete, parameters are adjusted using the difference in the equilibria of the two phases. In general, any CHL algorithm aims to adjust synaptic weight values leveraging conditioned iterative computation phases, with origins  grounded in Hopfield network-centric \cite{movellan1991contrastive}, Boltzmman-based learning \cite{hinton2002training}, and Helmholtz and neural heat exchange machines \cite{schmidhuber1989local,dayan1995helmholtz}. 

Like many bio-plausible alternatives to backprop, CHL schemes have been shown to approximate backprop as well as recurrent backprop \cite{scellier2019equivalence}. This approximation occurs as the strength of the nudging becomes infinitesimal, resulting the difference between the phases resembling a finite-difference gradient \cite{millidge2022backpropagation}. Other efforts have developed theory for EP, casting it in terms of a type of bilevel optimization \cite{zucchet2022beyond}. However, the original formulations of CHL schemes such as EP generally fail to scale up to complex tasks; to address this issue, variations of this CHL/EP have been designed that improve efficiency as well as performance on particular tasks \cite{laborieux2021scaling,scellier2018generalization,laydevant2021training}. In addition, \cite{laborieux2022holomorphic} defined the underlying energy function on the complex plane, which allows to avoid `nudging' via iterative computational steps, resulting in a more robust algorithm. 

\paragraph{Learning Dynamics.} 
For processes that adjust parameters values under the framework of contrastive Hebbian learning (CHL), the following energy functional serves as the central guide of the learning dynamics:
\begin{align}
    \mathcal{E}(\Theta) &= \sum^{L-1}_{\ell=1} \Big( \mathcal{C}(\mathbf{z}^\ell; \Theta_\ell) - \mathcal{C}(\mathbf{c}^\ell; \Theta_\ell) \Big) \label{eqn:chl_energy}
\end{align}
where $\mathbf{z}^\ell$ represents the incoming neural activity produced by running the neural system in a `positive' phase and $\mathbf{c}^\ell$ represents activity generated by running the system in a `negative' phase. The simplest possible local objective takes the form of Hopfield energy \cite{movellan1991contrastive}:
\begin{align}
    \mathcal{C}(\mathbf{z}^\ell;\Theta_\ell) = -(\mathbf{z}^{\ell-1})^{\mathsf{T}} \cdot \mathbf{W}^\ell \cdot \mathbf{z}^\ell.
\end{align}

To optimize the energy underlying a CHL system, gradients may be taken of the local Hopfield energy with respect to local parameters which results in a pair of two-term Hebbian rules: $\frac{\partial \mathcal{C}(\mathbf{z}^\ell,\mathbf{z}^{\ell-1};\Theta_\ell)}{\partial \Theta_\ell} = -\mathbf{z}^\ell \cdot (\mathbf{z}^{\ell-1})^T$ and  $\frac{\partial \mathcal{C}(\mathbf{c}^\ell,\mathbf{c}^{\ell-1};\Theta_\ell)}{\partial \Theta_\ell} = -\mathbf{c}^\ell \cdot (\mathbf{c}^{\ell-1})^T$. These result, when expressed in terms of gradient ascent, in synaptic updates in the form shown below:
\begin{align}
    \Delta \mathbf{W}^\ell = \Big(\mathbf{z}^\ell \cdot (\mathbf{z}^{\ell-1})^T\Big) - \Big(\mathbf{c}^\ell \cdot (\mathbf{c}^{\ell-1})^T\Big). \label{eqn:chl_update}
\end{align}
In order to obtain the necessary statistics for the local gradient terms for Equation \ref{eqn:chl_update}, a set of two conditioned dynamics is simulated in order to obtain the neural activity values required for computing local gradients. To do so, the equations that govern the system's dynamics are iteratively applied. These dynamics can be summarized in the following manner:
\begin{align}
    \mathbf{z}^\ell \leftarrow (1 - \gamma) \mathbf{z}^\ell + \gamma (\mathbf{W}^\ell \cdot \mathbf{z}^{\ell-1} + (\mathbf{W}^{\ell+1})^T \cdot \mathbf{z}^{\ell+1}) \label{eqn:layer_dynamics}
\end{align}
where we notice that neural activities change as a result of repeated local bottom-up and top-down synaptic transmission of nearby layer values. First, for each layer, Equation \ref{eqn:layer_dynamics} is applied $T$ times (after initializing activities), with the exception of the bottom and topmost layers -- these are clamped to sensory and context values, i.e., $\mathbf{z}^0 = \mathbf{x}$ and $\mathbf{z}^L = \mathbf{y}$, respectively -- to obtain the set of values $\{\mathbf{z}^{\ell,+}\}^{L}_{\ell=0}$ (positive phase statistics). Next, the topmost layer is then un-clamped and the same dynamics for each layer (still keeping the bottom layer fixed to sensory input) are further run for an additional $T$ steps to obtain $\{\mathbf{z}^{\ell,-}\}^{L}_{\ell=0}$ (negative phase statistics).

\paragraph{Equilibrium Propagation.} Despite being a general technique, that can be applied to other energy-based models, such as predictive coding, we have decided to include equilibrium propagation (EP) in the same section of CHL, as most of the literature of EP focuses on models that largely learn using the CHL energy of Equation~\ref{eqn:chl_energy} \cite{scellier2018generalization,scellier2023energy}. Generally speaking, however, EP consists in defining the energy term of the output layer $\mathcal L(\Theta_L)$ in terms of small local perturbations: first, an equilibrium is reached by minimizing the internal energy  $\mathcal{E}(\Theta)$ (hence, setting $\beta = 0$). Then, the output layer is \emph{nudged} towards the output signal by setting $\beta > 0$, but keeping is relatively small. At this point, learning happens as usual, where the energy to be minimized is a weighted combination $\mathcal{E}(\Theta) + \beta \mathcal{L}(\Theta_L)$.


\begin{table*}[!t]
\begin{center}
\scriptsize{
\begin{tabular}{|r||cccc|ccccc|c|c|c|} 
\hline

Method 
& \multicolumn{4}{c}{BP Constraints Resolved} 
& \multicolumn{5}{c}{New Capabilities} 
& Code/Software
& Spiking Impl. 
&  Hardware Impl.\\ [0.5ex] 

& WT & FL & BL & FBD 
& parallel & async. & 1-compute. & sparsity & no-diff 
&  &   & \\ [0.5ex] 

\hline\hline

PC 
& + & + & + &  
& - &  &  &  & +
&
\cite{ngclearn,pypc,predify,song2024inferring}

& \cite{rao2001spike,ororbia2019spiking} 
& 
\\

CHL 
&  &  + &  + &   
&  & & + &  & +
&
& \cite{neftci2014event,martin2021eqspike}

& 
\cite{zoppo2020equilibrium,oh2023memristor,foroushani2020analog}
\\
FO 
& + &  + &  + &  + 
&  + & + & + & + & + 
& \cite{Kohan_Signal_Propagation_The,pff_Ororbia}

& 
\cite{kohan2023signal,ororbia2023learning}

&
\cite{oguz2023forward}
\\ 

DFA 
& + &   &  + &   
& + &  &  & + &  
&
& \cite{samadi2017deep,zhao2020glsnn}

& \cite{filipovich2022silicon}
\\

TP 
& + &   &  & +  
& + &  & + &  & + 
& 
& 
&\\ 

LRA 
& + &   &  + &  
& - & - &  & + & + 
& 
& 
&\\ 
\hline
\end{tabular}
}

\end{center}
\caption{What issues of backprop does each biologically-plausible learning algorithm resolve and what new capabilities does it open? '+' = full or '-' = partial resolution of issue or availability of new capability (empty means no addressal of issue). 
}
\label{table:properties}
\end{table*}

\subsection{Forward-Only Learning} 
\label{sec:forward_only}

\paragraph{Background.} In recent years, a different set of biologically-plausible algorithms have emerged that avoid introducing or using feedback pathways that facilitate credit assignment. These have been often labeled as `forward-only' schemes \cite{kohan2018error,kohan2023signal} and generally construct a means of synaptic adjustment that relies on solely the forward inference process of a neural system. Some variations leverage the label as context information and run this through the feedforward propagation pathway \cite{kohan2018error,kohan2023signal} while others employ a mechanism for auto-generating adversarial `negative' data samples as context information that are then run through the feedforward pathway \cite{hinton2022forward,ororbia2023predictive} (i.e., such schemes minimize the ``goodness'' of adversarial data points and maximize goodness for those taken from the original dataset). 
Forward-only schemes have been generalized to spiking networks \cite{ororbia2023learning} and have also been demonstrated to work on neuromorphic chips \cite{wunderlich2021event}.

\paragraph{Learning Dynamics.} 
Although variations to the theme of forward-only (FO) exist, one may generically view any such scheme as optimizing the following functional:
\begin{align}
    \mathcal{E}(\Theta) &= \sum^{L-1}_{\ell=1} \mathcal{C}(\mathbf{z}^\ell, \mathbf{c}^\ell; \Theta_\ell)
    \label{eqn:ff_energy}
\end{align}
which is effectively a summation of local comparative or contrastive functions, depending on the variant of FO adaptation that is employed. For example, in signal propagation \cite{kohan2023signal}, one could employ a vector similarity measurement as follows:
\begin{align}
    \mathcal{C}(\mathbf{z}^\ell, \mathbf{c}^\ell; \Theta_\ell) = \frac{(\mathbf{c}^\ell)^{\mathsf{T}} \cdot \mathbf{z}^\ell}{||\mathbf{z}^\ell||_2 \mathbf{z}^\ell||_2}
\end{align}
or a contrastive measurement as in the forward-forward procedure \cite{hinton2022forward}:
\begin{align}
    \mathcal{C}(\mathbf{z}^\ell, \mathbf{c}^\ell; \Theta_\ell) = 
    \begin{cases}
        \log p(c=1; \mathbf{z}^\ell) & \text{if } \mathbf{x} \sim \mathcal{D} \\
        \log \big( 1 - p(c=1; \mathbf{c}^\ell) \big) & \text{if } \mathbf{x} \sim \mathcal{D}^{neg}
    \end{cases} \label{eqn:goodness_cost}
\end{align}
which is a simplified presentation of `goodness' in terms of a logistic regression objective. Notice that in Equation \ref{eqn:goodness_cost}, we make clear that the activity vector $\mathbf{z}^\ell$ is the product of sampling an input pattern from the original dataset $\mathcal{D}$ whereas $\mathbf{c}^\ell$ is sampled from a negative/adversarial data distribution $\mathcal{D}^{neg}$.

To optimize Equation \ref{eqn:ff_energy}, a reverse-mode differentiation may be used to calculate the partial derivatives of $\mathcal{L}(\Theta_L)$ and $\mathcal{C}(; \Theta_\ell)$ with respect to $\Theta_\ell$ for each layer (or block) $\ell$. These local gradients are then subsequently used to carry out a step of a variation of a first-order (or n-order) optimization process, such as Adam, to update parameters \cite{kohan2023signal,hinton2022forward}. Alternatively, a Hebbian learning update rule may be used to calculate the values needed for updating the parameters in tandem with an optimizer, such as Adam \cite{kohan2018error}.

\subsection{Other Learning Schemes}
\label{sec:other_algos}

Although this review focuses on three prominent bio-plausible frameworks, there is an expanding plethora of others being developed that have garnered the machine learning community's interest. 
Four key emerging approaches that we review here are direct feedback alignment (DFA) \cite{nokland2016direct}, target propagation (TP) \cite{lee2015difference,bengio2014auto}, local representation alignment (LRA) \cite{ororbia2019biologically}, and SoftHebb \cite{moraitis2022softhebb}. 

\paragraph{Direct Feedback Alignment.} Feedback alignment approaches focus on resolving the weight transport problem. 
An early effort, random feedback alignment \cite{lillicrap2016random}, used fixed random feedback projections to generate teaching signals for each layer (as opposed to the transpose of the feedforward weights).  DFA, on the other hand, introduced random feedback matrices that directly projected the output error to individual layers. It is worth noting that this method  has shown good performance on large-scale datasets \cite{moskovitz2018feedback,launay2020direct}, as well as promising implementation on photonic chips, given that it does not require layers to be updated sequentially during the backward pass. This enables DFA to be suitable for parallelization using photonics \cite{filipovich2022silicon}. Other variations incorporate mechanisms for learning the feedback matrices (instead of keeping them random/fixed) \cite{liao2016important}. 

\paragraph{Target Propagation.} Target propagation (TP) tackles the BP's problematic use of different operations for the forward and backward passes.  Instead of gradients, TP backpropagates target (vector) signals. Each pair of layers in a TP-trained network is treated as a shallow autoencoder where each layer tries to reconstruct the one below it, learning a local function and its (approximate) inverse. Notably, each autoencoder is designed with separate forward and backward connections and local rules are used to adjust their values. Recent theoretical work has improved the stability of TP and demonstrated its connection to Gauss-Newton optimization \cite{meulemans2020theoretical,bengio2020deriving}. 

\paragraph{Local Representation Alignment.} This bio-plausible algorithm can be viewed as a hybrid between TP and PC, engaging in a `coordinated local learning' process that minimizes a set of summed representation distance measurements \cite{ororbia2019biologically}. LRA-based approaches produce teaching signals by first adjusting layer-wise activities with an error modulated perturbation and then calculating local weight updates; notably, LRA can be recursively decomposed to produce asynchronous, parallel adjustments to portions of a network architecture, especially for larger convolutional neural systems \cite{ororbia2020large,zee2022robust,kappel2023block}. 

\paragraph{SoftHebb.} A noteworthy, recent line of work focuses on Hebbian adaptation, augmented with a soft winner-take-all activation function. In a fashion similar to PC, this algorithm can be shown to represent and learn a Bayesian generative model of the data \cite{moraitis2022softhebb}. When used for standard deep learning tasks, such as image classification, these algorithms have been shown to perform as well as other biologically plausible algorithms, also avoiding many of the implausibilities mentioned above \cite{journe2022hebbian}.

 \begin{table}[!t]
 \centering
 {
\setlength\tabcolsep{2pt}
\begin{tabular}{|l|c|}
 \hline
 \textbf{Domain} \& \textbf{Problem Solved} & \textbf{Method}\\
 \hline
 
 \specialcell[t]{l}{\textit{Physics}\\ \quad\quad  Does not require differentiation} & CHL, FO, LRA, TP  \\
  \specialcell[t]{l}{\textit{Hardware}\\ \quad\quad   Only needs inference architecture for learning} & FO   \\
 \specialcell[t]{l}{\textit{Optics/Physical Systems}\\ \quad\quad  No need to characterize complex physical systems} & FO, PC\\
 \specialcell[t]{l}{\textit{Physical Devices} \\ \quad\quad  Analog information processing} & FA, PC, FO \\
 \specialcell[t]{l}{\textit{Neuroscience} \\ \quad\quad  Fits complex neuronal models of learning} & PC, FO  \\
 \hline
\end{tabular}
}
\vspace{0.25cm}
\caption{A reference table for selecting the biologically-plausible learning method(s) that best fit(s) a specific (application) domain or solve a particular problem.}
\label{table:domain_to_method}
\end{table}

\section{Neuromorphic Systems}
\label{sec:neuromorphic}

Modern neuromorphic systems often aim to inscribe an ANN into an equivalent analog circuit \cite{davies2021advancing}. Currently, the best results are achieved by training a digital ANN on existing hardware and then transferring the synaptic weight values to the analog device. After transfer, there may be limited tuning of the ANN on the analog device. This limitation is primarily due to the lack of effective and scalable neuromimetic learning algorithms which can run natively in the analog domain, which many of the algorithms reviewed in this paper can do. The motivation of analog hardware is to bypass the escalating energy, computation, and financial burden of running ANNs on the digital intermediary layer. Instead, ANNs are implemented directly in the more efficient and scalable hardware layer. The eventual goal is to surpass the performance of digital artificial neural networks, not merely retain their performance at a lower cost. 

The current iteration of neuromorphic systems have much lower energy usage compared to equivalent digital hardware. However, they struggle to reach the growing level of digital ANN design complexity/scale -- key features accredited with the success of modern digital ANNs. Therefore, there are few applications where performance is comparable \cite{davies2021advancing}. This is due both to the ever-increasing size of state-of-the-art models, which cannot fit inside a single GPU, let alone on a single analog device, as well as the rapid pace of neural architecture development compared to hardware design.

These limitations of neuromorphic systems are due to constraints imposed by the architectural design of ANNs in hardware, when moving from the digital to analog domain \cite{neftci2017event}. However, many constraints faced by biological systems and those imposed by analog hardware are very similar. Moreover, the potential improvements in energy and compute efficiency of analog hardware is substantial. This naturally points towards a synergy between machine learning, neuroscience, and hardware design.

\section{Future Directions for Research}

While there has been significant progress and activity in recent years, biologically inspired (bio-inspired) learning methods have not yet reached the consistent and strong performance of backpropagation (backprop), which has been able to surpass human performance in multiple tasks. One of the grander goals in the field is to emulate the performance of backprop while retaining bio-plausible credit assignment. There are several research direction that would allow us to bridge this gap. Theoretically, we need to better understand the trajectory of learning and the final model state when learning with backprop as compared to bio-inspired credit assignment, and use this knowledge to understand which methods, or (the more likely) combination of methods, to focus future efforts on. In practice, empirical progress in the field will be obtained by adopting more of the methodology and techniques seen in machine learning research using backprop, including architecture design, regularization, and data processing. Most importantly, as this is an interdisciplinary field, potential breakthroughs will come if we manage to  stimulate the involvement of broader and diverse scientific communities. 

\subsection{Open Problems}

\paragraph{Flexible Libraries.} The first limitation to address in the field is the absence of a suitable software library (a notable exception is \cite{ngclearn}), similar to Pytorch or Tensorflow, in the sense that it provides flexibility and adaptability with different data sources and can use efficiently use varied computational resources such as GPUs, CPUs, TPUs for faster training. Furthermore, to encourage active participation in this domain from those who do not have  backgrounds in cognitive science or computational neuroscience/biophysics, there is a need to develop high-level APIs, similar in spirit to Keras, or design a plug-and-play-style environment.

\paragraph{Stability.} There is a need for theoretical understanding concerning convergence guarantees and the stability of bio-plausible approaches. Recent work on energy-based models, e.g., equilibrium propagation and predictive coding, are failing to generalize to very deep architectures, e.g., ResNets, and the best results available are limited to Alexnet or small VGGNets \cite{salvatori2022incremental,scellier2023energy}, models that are now about a decade old. A robust mathematical theory would provide essential insights into the conditions under which these alternative methods converge, their rate of convergence, and their stability across various network architectures and data distributions. This will be crucial for scaling these architectures to complex, deeper architectures as well as for applications in fields where reliability and precision are paramount, e.g., autonomous systems, medical diagnostics, and financial modeling. Such a theory would also guide the design of neural systems that are more resilient to the quality of initial parameters and training data, thereby enhancing the efficacy and efficiency of the training process.


\paragraph{Dynamics.} Implementations of such algorithms on edge devices will most likely need to handle dynamical environments. However, as of today, most research focuses on modelling static data. An invaluable direction of research would be to focus on time series data, such as next-frame prediction tasks. To this end, future efforts should focus on implementing bio-plausible update rules for deep learning models such as LSTMs, or for models inspired from control theory, such as Kalman filters. A successful example of such efforts, still on small-scale tasks, are works that use predictive coding to model active inference agents \cite{ororbia2023neuro}.

\subsection{Mortal Computation and Structural Evolution}
\label{sec:mortal_selection}
Many of the neuroscience-inspired credit assignment algorithms reviewed here contain components such as energy minimization that are well adapted to analog and neuromorphic hardware \cite{grollier2020neuromorphic,davies2021advancing}, yet are not necessarily competitive with backprop on digital hardware. This has inspired the possibility of designing specialized hardware for these algorithms which could theoretically obtain substantial improvements in speed, cost, and power-efficiency. At a hardware level, efficiency gains can only be achieved by utilizing the natural physics of the hardware to achieve the desired computation instead of `emulating' it digitally. The downside of this is that  computations are unavoidably affected by natural variability in the hardware introduced during the manufacturing process. While neural learning algorithms can route around these differences, this means that each network is intrinsically tied to and specialized for its physical hardware implementation in a way not true of purely digital programs. In effect, such neural computation is `mortal' \cite{ororbia2023mortal} since it cannot escape the destruction of its hardware substrate. Neuromimetic  algorithms hold the promise of being able to run effectively on such hardware, driving large gains to the efficiency of their models, at the cost of them becoming mortal.

In complement to the above point on structural substrate, we remark that investigating the combination and integration of biological learning schemes with forms of neuro-evolution, e.g., NeuroEvolution of Augmenting Topologies \cite{stanley2002evolving} (NEAT), and/or swarm intelligence \cite{eberhart2001swarm}, e.g., particle swarm optimization, could prove useful. Importantly, this line of inquiry would present a unique opportunity for constructing dynamical systems that automatically discover optimal network topologies for a given problem, potentially leading to more efficient neuronal structures overall. Furthermore, in theory, NEAT can adapt networks in dynamic environments, a trait beneficial to reinforcement learning and other areas where the problem space changes over time. 



\subsection{Application of AI in Science}
The increasing popularity of AI has motivated applications in various scientific areas such as physics \cite{raissi2019physics}, chemistry \cite{ji2021stiff}, health \cite{rajpurkar2022ai}, and biology, as well as on unconventional devices such as physical devices \cite{wright2022deep} and optics \cite{oguz2023forward}. Such computational devices, that rely on brain-like analog information processing, are still mostly based on backprop-based schemes that are unsuitable for physical implementation. In some cases, however, this is problematic. Physics-based neural network (PINNs), for example, suffer from two major drawbacks: first, they cannot find the actual solution when coupled with first-order optimization based on gradient descent, and need computationally expensive second-order quasi-newton optimization to work well; second, PINNs are still equipped with feed-forward connections and cannot efficiently capture temporal and spatial relationships due to difficulties in optimization and poor generalization guarantees \cite{raissi2019physics}; third, PINNs center around
 the necessity of having a differentiable activation function, which restricts their usage in most scientific areas and problems. On the other hand, bio-inspired alternatives overcome most of these issues. Approaches such as local representation alignment, predictive coding, and forward-only learning can work with non-differentiable activation functions \cite{kohan2018error,ororbia2019biologically}. Additionally, target propagation \cite{bengio2020deriving} and PC-based methods approximate quasi-Newton optimization, thus providing better gradients and are thus well-suited for physical implementation and applications in science.

\section{Discussion and Conclusion}
\label{sec:conclusion}

\paragraph{Impact in the Neurosciences.} 
Understanding the nature of credit assignment in the cortex is one of the fundamental goals in neuroscience and finding answers to the questions related to it would provide us with direct insight into many other fundamental questions related to brain function and organization. Deep learning, a close cousin of neuroscience, is now able to reach and surpass human-level performance on many tasks, solving problems similar in spirit to what the brain does using the backpropagation of errors algorithm, which is highly neuroscientifically implausible. However, machine learning problems and networks provide an invaluable test-bed to experiment with and to test the capabilities of more bio-plausible algorithms. While such experiments do not provide direct neuroscientific evidence, they can both test the capabilities of such algorithms (given that algorithms which cannot train even small neural systems are highly unlikely to be viable in the brain) as well as be used to quickly and easily test hypotheses in artificial neural networks which can then be verified (or not) by neuroscientific investigation. In the last decade, a number of significant algorithms, which are reviewed in this work, have been proposed as solutions to the problem of cortical credit assignment. While these can serve as starting points for development, there remain many open problems both in improving, scaling, and testing the capabilities of these algorithms, as well as experimentally checking the claims that they imply about cortical processing. 

\paragraph{Conclusion.} In this survey, we have described several important biologically-plausible algorithms for credit assignment in artificial neural networks, and discussed how they address several key drawbacks of backpropagation-based models. Despite not being yet prominently used in practical applications, progress in this direction would facilitate the training of deep neural networks using only local computations, a fundamental aspect of distributed computing and implementations on neuromorphic hardware. However, while this survey is primarily directed at computer scientists as well as machine learning engineers and practitioners, the field of neuromorphic computing is cross-disciplinary, as it touches fields like physics, material science, neuroscience, chemistry, and computer science \cite{mehonic2022brain}. By highlighting this interdisciplinarity, we aim to foster a more unified research landscape, enhancing accessibility for various disciplines to comprehend, improve, and apply these methods and frameworks effectively. 

To conclude, to ensure that progress goes in the right direction, it is important to reason about where such neuromimetic algorithms will be able to play a role in terms of real-world applications. Important examples include edge computing devices and autonomous neurorobotic systems, where energy efficiency is critical and computational parallelism is invaluable. Looking forward, identifying and targeting application-specific requirements will be key in steering the research related to biologically-inspired credit assignment algorithms towards solving real-world problems and challenges.

\small{
\bibliographystyle{acm}
\bibliography{ref}
}
\end{document}